\documentclass[letterpaper, 10 pt, conference]{ieeeconf}  

\IEEEoverridecommandlockouts                              

\usepackage{graphics} 
\usepackage{epsfig} 
\usepackage{mathptmx} 
\usepackage{times} 
\usepackage{amsmath} 
\usepackage{amssymb}  
\usepackage{xcolor}
\usepackage{algorithm}
\usepackage{algpseudocode}
\usepackage{booktabs}
\usepackage{hyperref}

\makeatletter
\renewcommand{\thetable}{\arabic{table}}
\renewcommand{\fnum@table}{Table~\thetable}
\let\paper@makecaption\@makecaption
\long\def\@makecaption#1#2{%
    \begingroup
    \ifx\@captype\@IEEEtablestring
        \let\@IEEEtablestring\relax
    \fi
    \paper@makecaption{#1}{#2}%
    \endgroup
}
\makeatother

\usepackage{tikz}
\usetikzlibrary{arrows.meta,calc,decorations.pathmorphing,backgrounds,shapes.geometric}

\definecolor{bodyDark}{HTML}{2F343B}
\definecolor{bodyMid}{HTML}{5A616B}
\definecolor{bodyLite}{HTML}{AEB6C0}
\definecolor{accent}{HTML}{2E7DD1}
\definecolor{accent2}{HTML}{E4933A}
\definecolor{groundcol}{HTML}{9AA0A8}

\title{\LARGE \bf
Bundled Contact Gradients: Stabilizing Differentiable Simulation for Deployable Dynamic Tasks
}

\author{Dyuman Aditya$^{1}$, Jin Cheng$^{2}$, Clemens Schwarke$^{2,3}$, \\ Quan Nguyen$^{1}$, Gaurav Sukhatme$^{1}$,  Stelian Coros$^{2}$, Gabriele Fadini $^{4}$  
\thanks{Corresponding author: Dyuman Aditya {\tt\small agdyuman@usc.edu}}
\thanks{$^{1}$University of Southern California, LA, USA.}%
\thanks{$^{2}$ETH Zurich, Switzerland.}%
\thanks{$^{3}$NVIDIA.}%
\thanks{$^{4}$ Zurich University of Applied Sciences, Switzerland.}%
}

\begin{document}

\maketitle
\thispagestyle{empty}
\pagestyle{empty}

\begin{abstract}

Differentiable simulation provides analytic gradients of robot dynamics, enabling fast and sample-efficient first-order policy optimization. However, obtaining smooth and informative gradients through rigid-body contact typically requires softened contact models, often at the expense of physical fidelity and thereby limiting learned policies largely to simulation. This trade-off becomes particularly consequential for dynamic humanoid motions, where accurate contact dynamics are critical for transferring policies to the real world. Increasing contact stiffness in rigid-body simulation improves the fidelity of interactions, but also makes the dynamics increasingly sensitive to small state perturbations, producing high-variance gradients that can destabilize first-order policy learning. To address this, we propose \emph{Bundled Contact Gradients (BCG)}, a contact-local randomized smoothing framework for differentiable policy learning. When stiff contact is detected, our method evaluates a local bundle of randomized perturbation rollouts around the stiff contact configuration and aggregates their gradient signal thereby reducing gradient variance. We demonstrate the effectiveness of our method by successfully training and transferring dynamic motions zero-shot onto a real-world Unitree G1 humanoid platform. Videos and supplementary information can be found at \href{https://bundledcontactgradients.github.io/}{\texttt{bundledcontactgradients.github.io}}

\end{abstract}

\section{INTRODUCTION}

Learning dynamic, contact-rich behaviors, such as agile locomotion and whole-body
motions is predominantly achieved by model-free reinforcement learning
(RL) \cite{hwangbo2019agile, peng2018deepmimic}. By optimizing policies without differentiating through the underlying dynamics, model-free RL can be applied directly to simulators with non-smooth contact models and has made it the workhorse behind recent sim-to-real successes in legged-robot control \cite{lee2020quadrupedal}. However, its reliance on zero-order gradient estimation makes policy optimization sample inefficient, particularly for long-horizon, high-dimensional control problems~\cite{xu2022shac}. As a result, training often requires billions of environment interactions and substantial wall-clock time~\cite{sferrazza2024humanoidbench}.

\begin{figure}[t]
    \centering
    \includegraphics[width=\columnwidth]{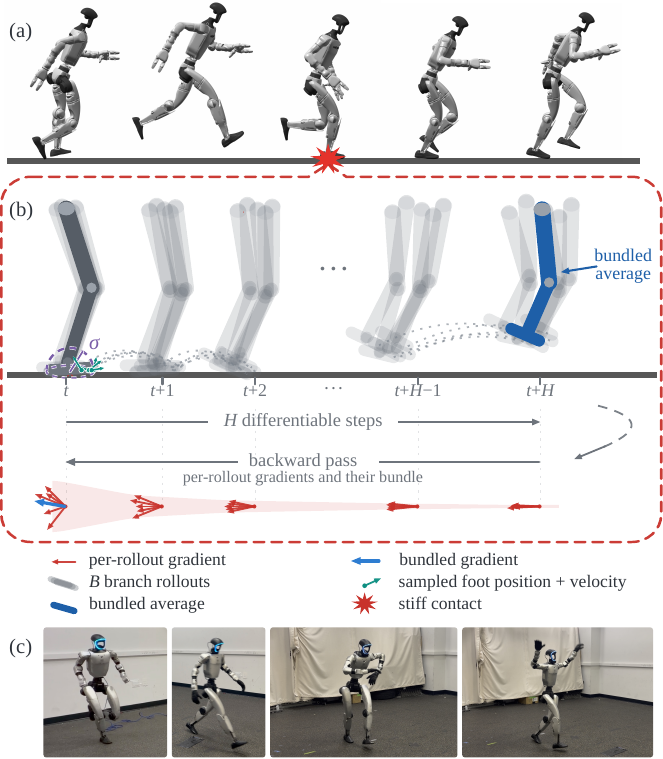}
    \caption{\textbf{Bundled Contact Gradients.}
    (a)~Dynamic, contact-rich Unitree G1 motions learned in differentiable
    simulation with stiff, deployable contact.
    (b)~At stiff contact (time $t$), $B$ perturbations of the contacting foot's
    position and velocity are sampled from a local sphere of radius $\sigma$
    and simulated in parallel for $H$ differentiable steps, then aggregated
    into the bundled state (blue). In the right-to-left backward pass,
    per-step sample gradients agree out of contact but diverge sharply at
    stiff contact (red). Bundling concentrates them around a common direction
    (blue), stabilizing first-order policy optimization at deployable contact
    stiffness.
    (c)~Zero-shot transfer of the learned policies to hardware.}
    \label{fig:teaser}
\end{figure}

Differentiable simulation offers a more efficient alternative. By exposing analytic
derivatives of the dynamics, it enables \emph{first-order} policy optimization, by backpropagating gradients through the simulated dynamics to the policy. When the dynamics
are smooth, the resulting gradients have far lower variance than their zero-order
counterparts, and methods such as SHAC~\cite{xu2022shac} report large gains in sample
and wall-clock efficiency over zero-order RL. The promise of first-order learning, however, has largely been confined to simulation due to highly softened contact models that are employed. 

Modeling stiff ground contacts in rigid-body differentiable simulation for legged control introduces a fundamental trade-off between physical fidelity and gradient quality. Compliant, low-stiffness contact yields smooth
gradients but permits soft, penetrating interactions that do not transfer to hardware.
Raising the stiffness toward physically faithful, deployable values restores realism
but makes the analytic gradient high-variance and biased across contact events~\cite{suh2022diffsim, metz2021gradients}, which
destabilizes first-order optimization.
Existing first-order methods largely sidestep this tension: they either soften contact
or shorten the differentiable horizon around contact events~\cite{georgiev2024ahac},
discarding gradient information where contact-rich behavior must be learned.

Our approach is to retain stiff contact dynamics while stabilizing the gradients
used for policy learning. At a stiff contact event, the gradient from a single
contact configuration can be highly sensitive to small perturbations; averaging
gradients over nearby configurations can provide a more stable learning signal.
Building on randomized smoothing and \emph{bundle
gradients}~\cite{DBLP:journals/corr/abs-2109-05143, LELIDEC2024101468}, we apply this
principle within first-order policy learning. But rather than applying smoothing globally, as in prior trajectory-optimization settings where evaluating bundles over entire trajectories can be computationally expensive, we apply it \emph{locally}, only at stiff contact events. We propose
\emph{Bundled Contact Gradients (BCG)} (Fig.~\ref{fig:teaser}): when stiff contact is detected, the simulator
evaluates a small bundle of perturbations sampled in Cartesian space, and simulated over a horizon in parallel. Aggregating
the branches yields a locally smoothed transition, so the forward rollout retains stiff,
deployable contact while the backward pass with automatic differentiation~\cite{macklin2022warp} receives a lower-variance, averaged contact
gradient. Because bundling is triggered only at stiff contact and executed in parallel on the
GPU, the overhead is limited and incurred only where the first-order gradient is most fragile.

We integrate BCG into Warp~\cite{macklin2022warp} and use it within a SHAC-style actor-critic 
framework for first-order policy optimization. To learn dynamic humanoid behaviors, we apply this framework to motion imitation, where the policy tracks reference motions using rewards from Adversarial Differential Discriminators (ADD)~\cite{zhang2025add}.

In summary, our main contributions are:
\begin{itemize}
    \item We introduce \textbf{Bundled Contact Gradients (BCG)}, a
    contact-local randomized-smoothing scheme that stabilizes first-order
    policy gradients under stiff contact without softening the forward dynamics.
    \item We show that BCG learns \textbf{deployable humanoid motion-tracking
    policies} with fewer samples and less training time than PPO, while
    achieving lower tracking error after sim-to-sim transfer.
    \item We demonstrate \textbf{zero-shot sim-to-real transfer} of dynamic
    motions learned through differentiable simulation to real humanoid hardware.
\end{itemize}

To the best of our knowledge, this work represents the first successful zero-shot transfer of a policy trained in differentiable simulation onto a real humanoid platform, as well as the first pairing of ADD with first-order differentiable-simulation policy learning.

\section{RELATED WORK}

\subsection{Reinforcement Learning for Dynamic Robot Control}
Model-free RL, and PPO~\cite{schulman2017ppo} in particular, is the dominant paradigm
for learning dynamic and contact-rich robot skills~\cite{handa2023dextreme, pmlr-v164-rudin22a}. Because it optimizes through
sampled returns rather than dynamics derivatives, it is robust to the discontinuous
physics of contact and underlies many sim-to-real results in legged locomotion~\cite{cheng2024parkour, doi:10.1126/scirobotics.adi9579}. However, this
robustness comes at the price of high-variance policy gradients and correspondingly poor sample efficiency, especially with higher dimensional systems like humanoids. 

\subsection{Differentiable Simulation and First-Order Policy Optimization}
Differentiable simulation has progressed from early engines~\cite{degrave2019differentiable,
deavila2018end} to GPU-accelerated frameworks~\cite{hu2020difftaichi, freeman2021brax,
macklin2022warp, geilinger2020add, qiao2021efficient}, enabling first-order policy
optimization by backpropagating through simulated
rollouts~\cite{mora2021pods, gillen2022leveraging, wiedemann2023apg}.
SHAC~\cite{xu2022shac} combines short differentiable windows with a learned critic,
and subsequent work extends first-order learning to deployable
locomotion~\cite{schwarke2025learningdeployablelocomotioncontrol, song2024quadruped},
agile flight~\cite{zhang2025learning, wiedemann2023apg}, and
multiphysics settings~\cite{xing2024stabilizing, amigo2025decoupled}.
For quadruped locomotion, Song et al.~\cite{song2024quadruped} combine forward simulation in a
non-differentiable simulator with gradients from a single rigid-body surrogate,
while Luo et al.~\cite{luo2024residual} guide SHAC with a frozen trotting policy
and learned residual actions. Schwarke et al.~\cite{schwarke2025learningdeployablelocomotioncontrol}
demonstrate zero-shot transfer of quadruped locomotion using an analytically
smoothed contact model, described in Section~\ref{subsec:contact-model}.
We build on this contact formulation to learn dynamic humanoid motions, where
smaller stability margins place greater demands on accurate and robust
foot--ground interactions. 

\subsection{Gradients through Stiff Contact}
The contact formulation in a differentiable simulator is important for both physical fidelity and gradient quality.

Soft penalty-based contact models ease differentiation by regularizing contact forces~\cite{freeman2021brax,geilinger2020add}, but excessive compliance can distort the foot--ground interactions needed for hardware transfer. Increasing the stiffness of these models improves physical fidelity, but makes the dynamics more difficult to resolve accurately with finite simulation timesteps, leading to inaccurate autodiff gradients~\cite{paulus2026differentiable}. To address this issue, Paulus et al.~\cite{paulus2026differentiable} introduce adaptive time integration to more accurately resolve stiff contact events and improve gradient fidelity. However, the resulting increase in simulation steps raises computational cost and lengthens the differentiable rollout which can lead to vanishing or exploding gradients.

Hard-contact models instead express non-penetration and friction through complementarity constraints, providing higher physical fidelity but less informative gradients around contact transitions. Werling et al.~\cite{werling2021fast} analytically differentiate a Linear Complementarity Problem (LCP). Dojo~\cite{howell2022dojo} uses a Nonlinear Complementarity Problem (NCP) and implicit differentiation of its interior-point solver. It obtains smooth contact gradients by implicitly differentiating a relaxed complementarity problem. However, neither reported implementation targets massively parallel GPU rollouts, and both can leave gradient-based optimization trapped in poor local solutions. 

Gradient propagation over long rollouts presents a further difficulty: repeated Jacobian products can cause vanishing or exploding gradients~\cite{metz2021gradients}. AHAC~\cite{georgiev2024ahac} adapts the differentiable horizon using a contact-stiffness constraint and bootstraps the remaining return with a critic. This limits exposure to unstable derivatives, but truncating at stiff contacts discards direct gradient information about how actions influence subsequent motion through these events, potentially removing sensitivities important for learning contact-rich behavior.

To combine informative gradients with physically faithful contact, we build on the analytically smoothed compliant contact model of Schwarke et al.~\cite{schwarke2025learningdeployablelocomotioncontrol}, described in Section~\ref{subsec:contact-model}, and increase its stiffness while stabilizing its gradients through local randomized smoothing.

\subsection{Randomized Smoothing and Bundled Gradients for Contact}
Randomized smoothing offers a way to obtain smoother contact responses by
averaging over nearby, randomly perturbed configurations. Bundled gradients replace the exact contact
derivative with a randomized-smoothing estimate and improve convergence of
trajectory optimization through contact~\cite{DBLP:journals/corr/abs-2109-05143}, an
approach that connects the sampling behavior of RL to analytic smoothing and enables
global planning over quasi-dynamic contact
models~\cite{pang2023global, pang2021convex}, complementing classical contact-implicit
trajectory optimization~\cite{posa2014direct, budhiraja2018differential}. Randomized
smoothing has likewise been used to augment differentiable physics and to control
nonsmooth systems~\cite{lidec2022augmenting, LELIDEC2024101468}, with recent work
addressing the bias such smoothing introduces~\cite{scheffler:hal-05629540}. These
methods operate largely in trajectory optimization or quasi-static manipulation and
smooth the contact model globally. In contrast, we adapt randomized smoothing to
first-order RL policy learning and apply it locally, only at stiff contacts, so that
dynamic, deployable motion-tracking can be trained under stiff contact.


\section{BACKGROUND}
\subsection{Reinforcement Learning}
\label{subsec:rl}

Policy learning is formalized as a Markov decision process
$(\mathcal{X}, \mathcal{A}, p, r, \gamma)$ with state $x_t \in \mathcal{X}$, action
$a_t \in \mathcal{A}$, transition density $p(x_{t+1} \mid x_t, a_t)$, reward
$r(x_t, a_t)$, and discount factor $\gamma \in [0,1)$. A stochastic policy
$\pi_{\theta}(a_t \mid x_t)$ with parameters $\theta$ induces
a trajectory distribution. The objective is to
maximize the expected discounted return
\[
    J(\theta) = \mathbb{E}_{\pi_{\theta}}\!\left[\sum_{t=0}^{N-1} \gamma^{t}\, r(x_t, a_t)\right].
\]
Model-free methods estimate $\nabla_{\theta} J$ using the likelihood-ratio
(score-function) identity
\[
    \nabla_{\theta} J
    =
    \mathbb{E}_{\pi_{\theta}}\!\left[\sum_{t=0}^{N-1}
    \nabla_{\theta} \log \pi_{\theta}(a_t \mid x_t)\, \hat{A}_t\right],
\]
where $\hat{A}_t$ is an advantage estimate~\cite{schulman2015gae}. This estimator
differentiates only the policy and treats the dynamics as a black box; it is therefore
a \emph{zero-order} gradient that requires no derivatives of the physics and is robust
to non-smooth contact, at the cost of variance that grows with the horizon and action
dimension. PPO~\cite{schulman2017ppo} is a widely used instance.

\subsection{Differentiable Simulation}
\label{subsec:diffsim}

A differentiable simulator replaces the black-box transition of
Section~\ref{subsec:rl} with a smooth, differentiable map
\[
    x_{t+1} = f_{\kappa}(x_t, a_t),
\]
where $\kappa$ parameterizes the stiffness of the contact model, and it exposes the
transition Jacobians $\partial f_{\kappa}/\partial x$ and
$\partial f_{\kappa}/\partial a$. The contact formulation used in this work is
described in Section~\ref{subsec:contact-model}. Rolling out a deterministic policy
$a_t = \pi_{\theta}(x_t)$ over a horizon $N$ produces a differentiable return
$J(\theta) = \sum_{t} \gamma^{t}\, r(x_t, a_t)$, whose gradient is obtained by
backpropagating through the unrolled computation graph, with the state sensitivity
propagated recursively through time,
\[
    \frac{\partial x_{t+1}}{\partial \theta}
    =
    \frac{\partial f_{\kappa}}{\partial x_t}\frac{\partial x_t}{\partial \theta}
    +
    \frac{\partial f_{\kappa}}{\partial a_t}\frac{\partial \pi_{\theta}}{\partial \theta}.
\]
This \emph{first-order} gradient is typically far lower in variance than the
score-function estimator, which is the source of the improved sample and wall-clock
efficiency of differentiable-simulation methods. The advantage holds only while
$f_{\kappa}$ is smooth: as $\kappa$ is increased toward physically faithful, deployable
contact, the Jacobians become stiff and the gradient develops large variance
across contact events. Overcoming this failure mode is the subject of
Section~\ref{sec:method}.

\begin{figure*}[!t]
    \centering
    \includegraphics[width=\textwidth]{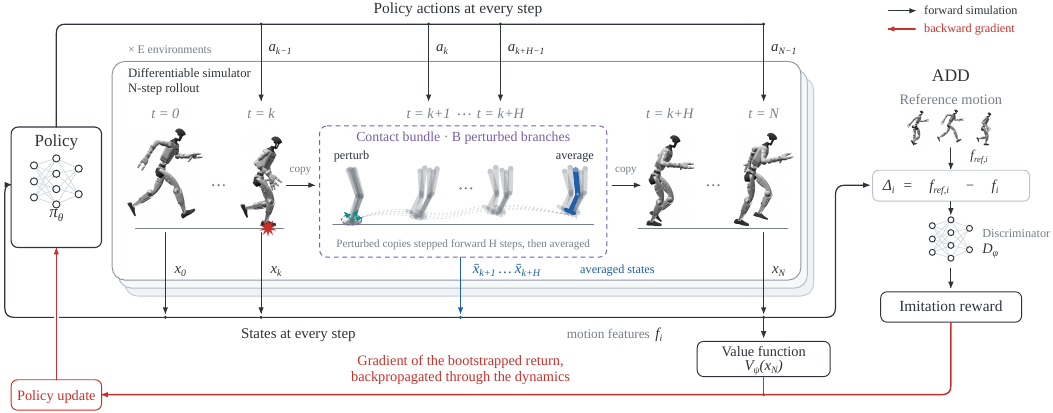}
    \caption{Policy learning with BCG in a SHAC-style framework and ADD.
    The policy drives parallel rollouts. At stiff contact, $B$ perturbed branches
    advance for $H$ steps under shared policy actions and are averaged to
    continue the rollout. ADD compares simulated and reference motion features
    to provide imitation rewards, whose gradients propagate through the
    dynamics to update the policy.}
    \label{fig:bcg-pipeline}
\end{figure*}

\subsection{Analytically Smoothed Contact Model}
\label{subsec:contact-model}

We use the contact model of Schwarke et
al.~\cite{schwarke2025learningdeployablelocomotioncontrol}, which replaces
discontinuous hard-contact activation with a continuous, analytically smoothed
response, enabling gradient computation through contact. Within Moreau's
time-stepping scheme~\cite{Moreau1988}, contact impulses are computed using a modified
Gauss--Seidel solver and scaled according to the penetration depth $d$ by
\[
    s_{\kappa}(d) = \frac{1}{1+\exp(-\kappa d)},
\]
where $d>0$ denotes penetration and $d<0$ denotes separation. This scaling 
replaces abrupt contact activation with
a smooth transition and permits forces at small separations, providing gradient
information before contact occurs.

The parameter $\kappa$ controls the extent of smoothing (i.e. the contact stiffness): smaller values produce
a wider, softer transition, whereas larger values reduce smoothing and approach
the original hard-contact response, increasing physical fidelity in the
rigid-contact regime. For finite $\kappa$, contact activation remains continuous,
although its derivatives become increasingly sensitive as $\kappa$ grows.
Throughout this paper, \emph{stiff contact} therefore means a large but finite
$\kappa$, rather than the discontinuous hard-contact limit.

BCG requires continuous contact dynamics with computable local derivatives,
since averaging derivatives cannot generally recover sensitivities across
discontinuous jumps. It therefore uses a large but finite $\kappa$ and local
randomized smoothing to stabilize gradients while remaining close to hard contact.



\subsection{First-Order Policy Optimization with SHAC}
\label{subsec:shac}

First-order policy optimization updates the policy using analytic gradients obtained by backpropagating through the differentiable rollout. While this can provide lower-variance updates than zero-order RL, differentiating through long horizons can cause vanishing or exploding gradients due to repeated Jacobian products~\cite{xu2022shac,georgiev2024ahac}. In this work we use Short Horizon Actor Critic (SHAC)~\cite{xu2022shac}, which collects short $N$-step differentiable rollouts from $E$ simulated environments in parallel and bootstraps the remaining return with a learned critic $V_{\psi}$. Its actor objective is
\[
    \mathcal{L}_{\mathrm{FO}}(\theta)
    =
    -\sum_{t=t_0}^{t_0+N-1} \gamma^{\,t-t_0}\, r(x_t, a_t)
    \;-\; \gamma^{\,N}\, V_{\psi}(x_{t_0+N}),
\]
with the critic trained by value regression. The short horizon limits unstable gradient propagation while retaining direct dynamics gradients over each rollout. SHAC nonetheless inherits the
contact sensitivity of the underlying gradient: at the high stiffness required for
deployable behavior, its truncated gradients can vary sharply with individual contact events, the
gap addressed in next Section~\ref{sec:method}.

\section{Bundled Contact Gradients}
\label{sec:method}

We propose \emph{Bundled Contact Gradients (BCG)} to stabilize first-order
policy learning through stiff contact. At a stiff contact event, BCG
creates nearby contact configurations, advances them in parallel, and
aggregates their states. Differentiating through this local bundle combines
sensitivities from several contact realizations, reducing reliance on any
single derivative. Each branch retains the large but finite contact
stiffness $\kappa$ of Section~\ref{subsec:contact-model}.
Figure~\ref{fig:bcg-pipeline} summarizes how BCG connects the simulator,
policy, and imitation objective.

Let $x_t$ be the simulator state at control step $t$, $a_t$ the action,
and $f_{\kappa}$ the differentiable transition between consecutive control
steps. We describe a bundle beginning at control step $k$, with
$B$ branches indexed by $i\in\{1,\ldots,B\}$ and duration $H\geq 1$ control steps.
Outside a bundle rollout, the
simulator follows an ordinary single branch transition.

BCG proceeds in three phases: contact-local perturbation, multi-step bundled
rollout, and gradient propagation. We describe these phases below, followed
by their integration into policy learning for motion imitation.

\subsection{Contact-Local Perturbations}
A bundle step is triggered when the set $\mathcal{C}_k$ of contacts whose normal
impulse exceeds a detection threshold $\tau$ is nonempty. We perturb the
contacting kinematic chain in Cartesian space so that the perturbation scale has a
direct physical interpretation. Let $q_k$ denote the joint positions of the contacting
kinematic chains and $p_{\mathcal{C}_k}(q_k)$ the current Cartesian positions
of their final links. For each branch $i$ We sample position and velocity
offsets and map them to joint-space increments:
\[
\begin{gathered}
    J_k = \frac{\partial p_{\mathcal{C}_k}}{\partial q}(q_k), \\[4pt]
    \begin{alignedat}{2}
        \Delta p^{(i)} &\sim \mathcal{N}(0,\sigma_p^2 I),
        \qquad & \Delta v^{(i)} &\sim \mathcal{N}(0,\sigma_v^2 I), \\
        \Delta q^{(i)} &= J_k^{\dagger}\Delta p^{(i)},
        & \Delta\dot q^{(i)} &= J_k^{\dagger}\Delta v^{(i)}.
    \end{alignedat}
\end{gathered}
\]
Here, $\mathcal{N}$ is a Gaussian distribution, $I$ is the identity matrix
of the Cartesian perturbation dimension, and $\sigma_p$ and $\sigma_v$
are the position and velocity standard deviations. $J_k^{\dagger}$ denotes
the damped pseudoinverse of the Jacobian $J_k$. Embedding the increments $\Delta q^{(i)}$ and
$\Delta\dot q^{(i)}$ into the simulator state gives an offset
$\delta x_k^{(i)}$ and initial branch state
$x_k^{(i)}=x_k+\delta x_k^{(i)}$; all other state components are unchanged.
During backpropagation, each offset $\delta x_k^{(i)}$ is treated as a
constant. Gradients therefore pass unchanged through the additive update
to $x_k$, without differentiating through perturbation sampling or its
Jacobian-based mapping.

\subsection{Multi-Step Bundled Rollout}
Starting from the perturbed states $x_k^{(i)}$, the branches evolve for
$H$ control steps under shared actions, initially using the current
action $a_k$ from the nominal rollout. Each transition $f_{\kappa}$ integrates
$S$ simulation substeps with the action held constant, after which the
branch states are averaged. Thus, even $H=1$ allows $S$ substeps of branch
evolution before the average supplies rewards and the next action.
Their propagation and aggregation are
\begin{equation}
\label{eq:bcg-rollout}
\begin{aligned}
    x_{t+1}^{(i)} &= f_{\kappa}(x_t^{(i)},a_t), \\
    \bar{x}_{t+1} &= \mathcal{A}\big(x_{t+1}^{(1)},\ldots,x_{t+1}^{(B)}\big),
    \qquad k\leq t<k+H,
\end{aligned}
\end{equation}
where $\mathcal{A}$ aggregates the branch states into a representative
state $\bar{x}_{t+1}$, using arithmetic means for positions and velocities.

At subsequent control boundaries ($t>k$), the policy with parameters
$\theta$ computes $a_t=\pi_{\theta}(\bar{x}_t)$, and the same action is
applied to every branch. The averaged state also supplies the observation
and reward at that boundary. The nominal rollout then resumes from
$x_{k+H}=\bar{x}_{k+H}$. Algorithm~\ref{alg:bundled_contacts_step}
summarizes this procedure for one environment. However, in our implementation branch transitions over \textit{all} bundling environments run in parallel on the GPU to reduce overhead.

\begin{algorithm}[t]
\caption{BCG rollout segment with duration $H$ control steps}
\label{alg:bundled_contacts_step}
\begin{algorithmic}[1]
\Require State $x_k$, current action $a_k$, policy $\pi_\theta$, simulator $f_\kappa$ ($S$ substeps per transition), bundle size $B$, duration $H$, scales $\sigma_p,\sigma_v$, threshold $\tau$
\State Detect contacts $\mathcal{C}_k$ using threshold $\tau$
\If{$\mathcal{C}_k=\emptyset$}
    \State \Return $f_\kappa(x_k,a_k)$ at step $k+1$
\EndIf
\State Compute Jacobian $J_k$ and $J_k^\dagger$
\For{$i=1,\ldots,B$ \textbf{in parallel}}
    \State Sample $\Delta p^{(i)},\Delta v^{(i)}$ using $\sigma_p,\sigma_v$
    \State Map to $\Delta q^{(i)},\Delta\dot q^{(i)}$ using $J_k^\dagger$
    \State Initialize $x_k^{(i)}\gets x_k+\delta x_k^{(i)}$
\EndFor
\For{$t=k,\ldots,k+H-1$}
    \If{$t>k$}
        \State $a_t\gets\pi_\theta(\bar{x}_t)$
    \EndIf
    \For{$i=1,\ldots,B$ \textbf{in parallel}}
        \State $x_{t+1}^{(i)}\gets f_\kappa(x_t^{(i)},a_t)$
    \EndFor
    \State $\bar{x}_{t+1}\gets\mathcal{A}(x_{t+1}^{(1)},\ldots,x_{t+1}^{(B)})$
\EndFor
\State \Return $\bar{x}_{k+H}$ at step $k+H$
\end{algorithmic}
\end{algorithm}

\subsection{Gradient Propagation}

Let $Z_t=dx_t/d\theta$ and $U_t=da_t/d\theta$ denote the total
sensitivities of the state and action to the policy parameters.
Write $F_{x,t}=\partial f_\kappa/\partial x$ and
$F_{a,t}=\partial f_\kappa/\partial a$ for the transition Jacobians,
evaluated at $(x_t,a_t)$. Without bundling, gradients propagate along
a single trajectory:
\begin{equation}
\label{eq:single-gradient}
    Z_{t+1}=F_{x,t}Z_t+F_{a,t}U_t.
\end{equation}
BCG instead propagates a sensitivity $Z_t^{(i)}=dx_t^{(i)}/d\theta$
along each perturbed branch, with Jacobians evaluated at
$(x_t^{(i)},a_t)$. Since the perturbations are treated as constants,
$Z_k^{(i)}=Z_k$. For $k\leq t<k+H$,
\begin{equation}
\label{eq:bcg-gradient}
\begin{aligned}
    Z_{t+1}^{(i)} &= F_{x,t}^{(i)}Z_t^{(i)}+F_{a,t}^{(i)}U_t, \\
    \bar{Z}_{t+1} &=
    \sum_{i=1}^{B}
    \frac{\partial\mathcal{A}}{\partial x_{t+1}^{(i)}}Z_{t+1}^{(i)},
\end{aligned}
\end{equation}
where $\bar{Z}_{t+1}=d\bar{x}_{t+1}/d\theta$. For state components
aggregated arithmetically, the corresponding rows of this expression become
\begin{equation}
\label{eq:bcg-gradient-mean}
    \bar{Z}_{t+1}
    =\underbrace{\frac{1}{B}\sum_{i=1}^{B}
    \left(F_{x,t}^{(i)}Z_t^{(i)}+F_{a,t}^{(i)}U_t\right)}
    _{\text{averaging sensitivities across perturbed branches}}.
\end{equation}
This average is the local smoothing mechanism: it can attenuate sharp,
contact-dependent variations by combining sensitivities from nearby
trajectories.
The shared action sensitivity $U_t$ includes the policy's dependence on
$\bar{x}_t$ for $t>k$, so the recurrence retains feedback through all
$H \cdot S$ simulation substeps. These state and action sensitivities enter the
imitation-objective gradient through the chain rule, as described next.

\subsection{Policy Learning and Imitation Objective}
We use this bundled rollout inside SHAC~\cite{xu2022shac}, minimizing
$\mathcal{L}_{\mathrm{FO}}$ from Section~\ref{subsec:shac} with rewards
and critic inputs evaluated on the averaged states during a bundle.

To learn dynamic humanoid behaviors, we apply this framework to motion
imitation using Adversarial Differential Discriminators
(ADD)~\cite{zhang2025add}. We choose ADD because its differentiable reward
allows backpropagation through both the objective and simulated rollouts,
while its learned aggregation avoids the manual tuning of individual
tracking-reward weights used in other motion-imitation frameworks.
The imitation reward is
\begin{equation}
\label{eq:add-reward}
\begin{aligned}
    r_t^{\mathrm{ADD}} &= -\log\!\big(1-D_{\varphi}(\Delta_t)\big), \\
    \Delta_t &= f_{\mathrm{ref},t}-f_t,
\end{aligned}
\end{equation}
where $f_t$ and $f_{\mathrm{ref},t}$ are simulated and phase-matched reference
motion features. The discriminator $D_{\varphi}$ is trained to distinguish
the ideal zero residual from policy-generated residuals, providing an adaptive
tracking objective. We use this reward in $\mathcal{L}_{\mathrm{FO}}$ and
backpropagate through the discriminator's input and the bundled simulator
rollout to update the policy. ADD thus supplies the imitation objective,
while BCG stabilizes its propagation through stiff contact.

\section{EXPERIMENTS}
\label{sec:experiments}

Our experiments are designed to test whether Bundled Contact Gradients
(BCG) enable first-order policy optimization under \emph{stiff, deployable}
contact, and if the resulting policies transfer to another simulator and hardware. We
organize the evaluation around five questions: \textbf{1)} Is stiff contact necessary for dynamic humanoid motions? (Sec.~\ref{subsec:exp-softfail}),
\textbf{2)} How does BCG reduce gradient variance under stiff contact and stabilize gradient propagation? (Sec.~\ref{subsec:exp-gradients}),
\textbf{3)} How do SHAC+BCG, vanilla SHAC, and PPO compare in sample efficiency, training time, and convergence? (Sec.~\ref{subsec:exp-efficiency}),
\textbf{4)} How does SHAC+BCG's tracking accuracy compare with PPO after sim-to-sim transfer?  (Sec.~\ref{subsec:exp-tracking})
\textbf{5)} Do the learned policies transfer zero-shot to real humanoid hardware? (Sec.~\ref{subsec:exp-sim2real}).

\subsection{Experimental Setup}
\label{subsec:exp-setup}
We evaluate our framework on the Unitree G1 humanoid by training policies
to imitate four dynamic, contact-rich motions from the LAFAN1
dataset~\cite{harvey2020robust}: Run, Jump, Fight, and Dance, each lasting
15\,s. Policies are trained in the differentiable rigid-body simulator of~\cite{schwarke2025learningdeployablelocomotioncontrol},
which implements an analytically smoothed contact model in NVIDIA
Warp~\cite{macklin2022warp}. We assess sim-to-sim transfer by evaluating
the trained policies in MuJoCo. All experiments are run on an NVIDIA
RTX~2080~Ti GPU. All SHAC+BCG experiments use a differentiable rollout horizon of $N=32$ control steps, bundle size $B=10$, bundle duration $H=2$, and Cartesian perturbation scales $\sigma_p=1$ cm and $\sigma_v=2$ cm/s. We define stiff contact as a normal contact force exceeding $\tau=400$ N (which is greater than the robot's body weight), that triggers bundling. Other training parameters are listed on the website.

\subsection{Results}
\begin{table}[t]
    \centering
    \footnotesize
    \setlength{\tabcolsep}{2.5pt}
    \begin{tabular}{@{}ccccccccc@{}}
        \toprule
        & \multicolumn{2}{c}{Run}
        & \multicolumn{2}{c}{Jump}
        & \multicolumn{2}{c}{Fight}
        & \multicolumn{2}{c}{Dance} \\
        \cmidrule(lr){2-3}\cmidrule(lr){4-5}
        \cmidrule(lr){6-7}\cmidrule(l){8-9}
        $\kappa$ & Falls $\downarrow$ & $\Delta d\downarrow$
                 & Falls $\downarrow$ & $\Delta d\downarrow$
                 & Falls $\downarrow$ & $\Delta d\downarrow$
                 & Falls $\downarrow$ & $\Delta d\downarrow$ \\
        \midrule
        50  & 5/5 & 10.6 & 5/5 & 12.6 & 5/5 & 13.3 & 5/5 & 13.5 \\
        100 & 5/5 & 8.1  & 4/5 & 10.8 & 3/5 & 9.1  & 4/5 & 9.5 \\
        \textbf{300}
            & \textbf{0/5} & \textbf{3.1}
            & \textbf{0/5} & \textbf{3.3}
            & \textbf{0/5} & \textbf{3.4}
            & \textbf{0/5} & \textbf{2.9} \\
        \bottomrule
    \end{tabular}
    \caption{\textbf{Contact-stiffness ablation with BCG (Q1).}
    All policies are trained with SHAC+BCG at the indicated stiffness
    $\kappa$ and evaluated over 5~runs after transfer to MuJoCo.
    Falls: number of runs in which root height drops below 0.3\,m,
    reported as $n/5$.
    $\Delta d$: difference in mean foot penetration depth [mm] between the
    training simulator and MuJoCo; positive values indicate deeper
    penetration during training.}
    \label{tab:q1_transfer}
\end{table}

\subsubsection{Q1: Effect of Contact Stiffness on Transfer}
\label{subsec:exp-softfail}
We vary the training contact stiffness $\kappa$ while using the same
SHAC+BCG training procedure at every stiffness level, including the soft
contact settings. Table~\ref{tab:q1_transfer} reports falls after transfer
to MuJoCo and the difference in foot penetration between the training
simulator and MuJoCo. As stiffness $\kappa$ is raised, the number of falls goes down showing how contact fidelity in differentiable simulation affects transfer reliability.

\begin{figure}[!t]
    \centering
    \includegraphics[width=\columnwidth,trim=0 6bp 0 2bp,clip]{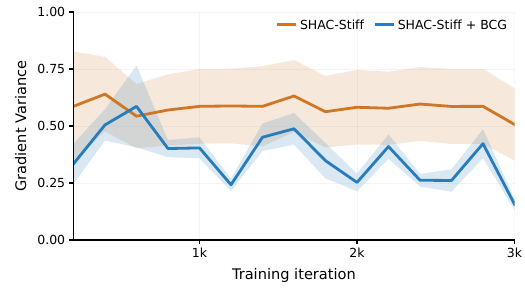}
    \caption{\textbf{Policy-gradient variance during training (Q2).} 
    Shading shows $\pm$ one standard deviation across 4,000 resamples of the 128 training environments, showing how much the estimated variance changes when the recorded gradients are resampled.}
    \label{fig:gradient_diagnostics}

    \par\vspace{4pt}
    \centering
    \includegraphics[width=\columnwidth]{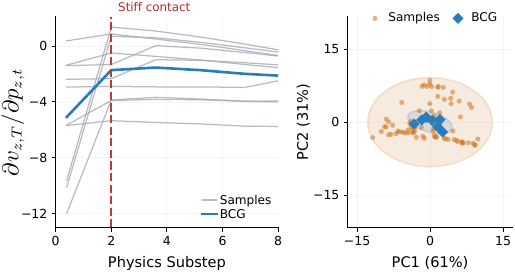}
    \par\vspace{1pt}
    \makebox[\columnwidth][l]{%
        \hspace{0.330\columnwidth}\makebox[0pt]{\footnotesize (a)}%
        \hspace{0.507\columnwidth}\makebox[0pt]{\footnotesize (b)}}
    \caption{\textbf{Contact-local gradient averaging Q2.}
    (a) One contact bundle with 10 samples during stiff contact over a window of $H=2$ (4 physics substeps per control step)
    (b) Ellipses summarize the spread of the gradient point clouds, where orange points are gradients from individual perturbed contact samples and blue points are the corresponding BCG-averaged gradients. Each ellipse encloses 95\% of the probability mass of a fitted Gaussian distribution.}
    \label{fig:contact_gradients}
\end{figure}

\subsubsection{Q2: Gradient Variance and Contact Sensitivities}
\label{subsec:exp-gradients}
Retaining stiff contact requires gradients that remain
stable enough during training. We therefore examine gradient variability
during training with and without BCG as well as understand how BCG works to reduce gradient variance at a low level.
For the Jump motion, we compare policy-gradient variance across all training environments at
saved checkpoints.
We compute variance by summing the sample variances of all
policy-gradient components across environments.
Figure~\ref{fig:gradient_diagnostics} shows lower variance with BCG, indicating greater agreement among the environment-wise gradients combined in each policy update. In other words, fewer environments produce conflicting parameter-update directions, resulting in a more consistent learning signal.

To examine the underlying mechanism in BCG, we rollout $N=32$ steps of the differentiable simulator with a single environment ($\kappa=300$) and measure how small changes in the state affect the G1's pelvis' vertical velocity within a bundle stage.
Figure~\ref{fig:contact_gradients}(a) plots one element of the state-gradient vector: $\partial v_{z,T}/\partial p_{z,t}$, which measures the sensitivity of future pelvis vertical velocity to the current pelvis height over the bundle window. We choose this component because vertical pelvis motion in the Jump task is strongly coupled to foot--ground contact, making it an intuitive measure of contact-induced gradient sensitivity.
At stiff contact, individual perturbed branches exhibit markedly
different sensitivities, while their bundled average attenuates the
extreme responses. This illustrates how BCG reduces dependence on a
single contact realization that would otherwise cause varied gradients, while retaining stiff dynamics.

Figure~\ref{fig:contact_gradients}(b) examines the sensitivity of the
same pelvis vertical velocity to all input-state components
over the bundle window. We collect these sensitivities into vectors and
use Principal Component Analysis (PCA) to display them in a shared
two-dimensional view. Each orange point represents one perturbed branch,
and each blue diamond represents the bundled gradient from one sampling
draw. Nearby points have similar sensitivity patterns in this projection.
The orange points are widely spread, whereas the blue points cluster
closely together. The smaller blue ellipse summarizes this reduction in
variation across bundle draws, showing that the averaging effect extends
beyond pelvis height to the joint variation of the full state-gradient
vector.

Together, these results show how contact-local averaging reduces
variability in simulator sensitivities and provides a more consistent
learning signal under stiff contact. By moderating contact-induced
gradient fluctuations, BCG helps keep gradients stable over longer
rollout horizons compared to previous work~\cite{schwarke2025learningdeployablelocomotioncontrol}.

\begin{figure}[t]
    \centering
    \includegraphics[width=\columnwidth]{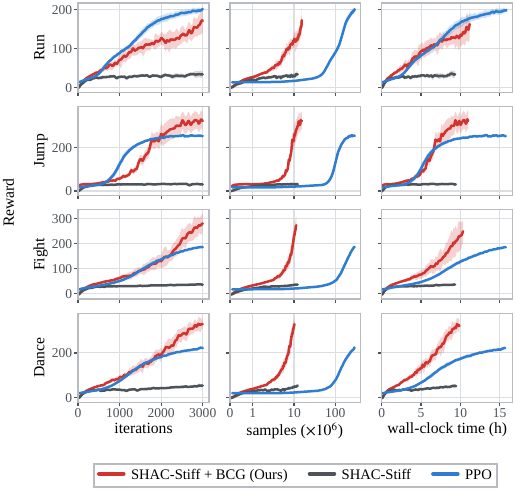}
    \caption{Learning curves on the four reference motions (Rows) plotted against
    three budgets: training iterations (left), environment samples (middle),
    and wall-clock time (right). The sample axis is
    logarithmic}
    \label{fig:sample_efficiency}
\end{figure}

\subsubsection{Q3: Learning Efficiency and Convergence}
\label{subsec:exp-efficiency}
Figure~\ref{fig:sample_efficiency} shows SHAC+BCG reaching comparable
reward levels with over an order of magnitude fewer samples than PPO,
while vanilla SHAC stalls at low reward. BCG reaches higher final
rewards compared to PPO in a few motions as well.

SHAC+BCG and vanilla SHAC consume comparable sample budgets at a given
iteration count. BCG's more informative gradients allow training with
64 environments instead of 128, offsetting the additional branch samples
used for bundling at stiff contacts. BCG therefore learns substantially
more from a similar simulation budget. BCG does however have a slightly higher computational overhead compared with Vanilla SHAC due to a larger computational graph for backpropagation.

\begin{table}[t]
    \centering
    \small
    \setlength{\tabcolsep}{4pt}
    \begin{tabular*}{\columnwidth}{@{\extracolsep{\fill}}lcc@{}}
        \toprule
        Motion & SHAC-Stiff + BCG [cm] & PPO [cm] \\
        \midrule
        Run   & \textbf{34.1 $\pm$ 0.4} & 42.8 $\pm$ 0.7 \\
        Jump  & 18.6 $\pm$ 0.4 & \textbf{18.0 $\pm$ 0.2} \\
        Fight & \textbf{13.5 $\pm$ 0.3} & 34.3 $\pm$ 1.4 \\
        Dance & \textbf{18.5 $\pm$ 0.8} & 19.0 $\pm$ 0.1 \\
        \bottomrule
    \end{tabular*}
    \caption{\textbf{Tracking accuracy after sim-to-sim transfer (Q4).}
    Global mean per-body position error $E_{\mathrm{track}}$ [cm],
    reported as mean $\pm$ std over 5~runs.}
    \label{tab:tracking_main}
\end{table}

\subsubsection{Q4: Tracking Performance across Methods}
\label{subsec:exp-tracking}
After transfer to MuJoCo, we compare SHAC+BCG and PPO using global
mean per-body position error (Table~\ref{tab:tracking_main}).
SHAC+BCG achieves lower mean error on three of the four motions.
PPO has slightly lower error on Jump, but its policy keeps both feet
on the ground instead of reproducing the jumping motion, as shown in
the accompanying videos. These results highlight the potential of first-order learning with informative gradients, which can guide the policy toward more faithful motion tracking.

\subsubsection{Q5: Sim-to-Real Transfer}
\label{subsec:exp-sim2real}
Finally, we deploy the learned SHAC+BCG policies zero-shot on the real Unitree G1 without any hardware fine-tuning. The robot successfully executes the learned dynamic motions, providing qualitative evidence that the stiff-contact policies transfer beyond simulation; videos of the hardware results are available on the project website.

\label{subsec:exp-sim2real}


\section{CONCLUSION}
We presented \emph{Bundled Contact Gradients (BCG)}, a contact-local randomized smoothing method for stabilizing first-order policy learning under stiff, deployable contact dynamics. BCG aggregates sensitivities across nearby contact configurations, providing a more consistent learning signal while retaining the contact fidelity required for transfer. Integrated with SHAC and ADD, BCG enables dynamic humanoid motion learning with over an order of magnitude fewer environment samples than PPO, improves tracking accuracy after transfer to MuJoCo compared to PPO, and enables zero-shot deployment of the learned motions on a real Unitree G1. These results demonstrate that the analytic gradients provided by differentiable simulation can remain useful even in the stiff contact regime required for real-world dynamic control.

BCG introduces additional computation and memory from simulating and differentiating through multiple contact branches. While contact-local activation limits this cost, tasks with sustained stiff contact may incur greater overhead than PPO. The perturbation scales and number of bundle samples also require tuning to balance smoothing, bias, and computational cost. Future work could adapt these parameters automatically based on local contact sensitivity and extend BCG to higher-dimensional, contact-rich problems such as humanoid loco-manipulation and dexterous manipulation. More broadly, we view BCG as a step toward making differentiable simulation a practical tool for learning increasingly complex, deployable robotic behaviors efficiently.

\bibliographystyle{IEEEtran}
\bibliography{bibliography}

@article{schulman2017ppo,
  title = {Proximal Policy Optimization Algorithms},
  author = {Schulman, John and Wolski, Filip and Dhariwal, Prafulla and Radford, Alec and Klimov, Oleg},
  journal = {arXiv:1707.06347},
  year = {2017},
}

@inproceedings{schulman2015gae,
  title = {High-Dimensional Continuous Control Using Generalized Advantage Estimation},
  author = {Schulman, John and Moritz, Philipp and Levine, Sergey and Jordan, Michael and Abbeel, Pieter},
  booktitle = {International Conference on Learning Representations},
  year = {2016},
}

@inproceedings{hu2020difftaichi,
  title = {DiffTaichi: Differentiable Programming for Physical Simulation},
  author = {Hu, Yuanming and Anderson, Luke and Li, Tzu-Mao and Sun, Qi and Carr, Nathan and Ragan-Kelley, Jonathan and Durand, Fr{\'e}do},
  booktitle = {International Conference on Learning Representations},
  year = {2020},
}

@article{freeman2021brax,
  title = {Brax -- A Differentiable Physics Engine for Large Scale Rigid Body Simulation},
  author = {Freeman, C. Daniel and Frey, Erik and Raichuk, Anton and Girgin, Sertan and Mordatch, Igor and Bachem, Olivier},
  journal = {arXiv:2106.13281},
  year = {2021},
}

@misc{macklin2022warp,
  title={Warp: A High-Performance Python Framework for GPU Simulation and Graphics},
  author={Macklin, Miles},
  year={2022},
  howpublished={NVIDIA GPU Technology Conference (GTC)},
  note={\url{https://github.com/NVIDIA/warp}}
}

@inproceedings{xu2022shac,
  title = {Accelerated Policy Learning with Parallel Differentiable Simulation},
  author = {Xu, Jie and Makoviychuk, Viktor and Narang, Yashraj and Ramos, Fabio and Matusik, Wojciech and Garg, Animesh and Macklin, Miles},
  booktitle = {International Conference on Learning Representations},
  year = {2022},
}

@inproceedings{georgiev2024ahac,
  title = {Adaptive Horizon Actor-Critic for Policy Learning in Contact-Rich Differentiable Simulation},
  author = {Georgiev, Ignat and Srinivasan, Krishnan and Xu, Jie and Heiden, Eric and Garg, Animesh},
  booktitle = {International Conference on Machine Learning},
  year = {2024},
}

@inproceedings{suh2022diffsim,
  title = {Do Differentiable Simulators Give Better Policy Gradients?},
  author = {Suh, Hyung Ju Terry and Simchowitz, Max and Zhang, Kaiqing and Tedrake, Russ},
  booktitle = {International Conference on Machine Learning},
  year = {2022},
}

@article{metz2021gradients,
  title = {Gradients are Not All You Need},
  author = {Metz, Luke and Freeman, C. Daniel and Schoenholz, Samuel S. and Kachman, Tal},
  journal = {arXiv:2111.05803},
  year = {2021},
}

@inproceedings{zhang2025add,
  title={Physics-Based Motion Imitation with Adversarial Differential Discriminators},
  author={Zhang, Ziyu and Bashkirov, Sergey and Yang, Dun and Shi, Yi and Taylor, Michael and Peng, Xue Bin},
  booktitle={SIGGRAPH Asia 2025 Conference Papers},
  year={2025},
  doi={10.1145/3757377.3763819}
}

@article{xing2024stabilizing,
  title = {Stabilizing Reinforcement Learning in Differentiable Multiphysics Simulation},
  author = {Eliot Xing and Vernon Luk and Jean Oh},
  journal = {International Conference on Learning Representations},
  year = {2025},
}

@inproceedings{paulus2026differentiable,
  title = {Differentiable Simulation of Hard Contacts with Soft Gradients for Learning and Control},
  author = {Anselm Paulus and Andreas Ren{\'e} Geist and Pierre Schumacher and V{\'\i}t Musil and Simon Rappenecker and Georg Martius},
  booktitle = {International Conference on Learning Representations},
  year = {2026},
}

@unpublished{scheffler:hal-05629540,
  TITLE = {{Unbiased First-Order Randomized Smoothing for Differentiable Simulators}},
  AUTHOR = {Scheffler, Mathis and Jallet, Wilson and Schmid, Cordelia and Carpentier, Justin},
  URL = {https://hal.science/hal-05629540},
  NOTE = {Preprint},
  YEAR = {2026},
}

@article{LELIDEC2024101468,
  title = {Leveraging randomized smoothing for optimal control of nonsmooth dynamical systems},
  journal = {Nonlinear Analysis: Hybrid Systems},
  volume = {52},
  pages = {101468},
  year = {2024},
  doi = {https://doi.org/10.1016/j.nahs.2024.101468},
  author = {Quentin {Le Lidec} and Fabian Schramm and Louis Montaut and Cordelia Schmid and Ivan Laptev and Justin Carpentier},
}

@article{zhang2025learning,
  title={Learning vision-based agile flight via differentiable physics},
  author={Zhang, Yuang and Hu, Yu and Song, Yunlong and Zou, Danping and Lin, Weiyao},
  journal={Nature Machine Intelligence},
  pages={1--13},
  year={2025},
  publisher={Nature Publishing Group}
}

@article{DBLP:journals/corr/abs-2109-05143,
  author = {H. J. Terry Suh and
                  Tao Pang and
                  Russ Tedrake},
  title = {Bundled Gradients through Contact via Randomized Smoothing},
  journal = {CoRR},
  volume = {abs/2109.05143},
  year = {2021},
}

@article{pang2023global,
  title={Global Planning for Contact-Rich Manipulation via Local Smoothing of Quasi-Dynamic Contact Models},
  author={Pang, Tao and Suh, H. J. Terry and Yang, Lujie and Tedrake, Russ},
  journal={IEEE Transactions on Robotics},
  volume={39},
  number={6},
  pages={4691--4711},
  year={2023},
  doi={10.1109/TRO.2023.3300230}
}

@article{lidec2022augmenting,
  title = {Augmenting Differentiable Physics with Randomized Smoothing},
  author = {Le Lidec, Quentin and Montaut, Louis and Schmid, Cordelia and Laptev, Ivan and Carpentier, Justin},
  journal = {arXiv:2206.11884},
  year = {2022},
}

@article{posa2014direct,
  title={A direct method for trajectory optimization of rigid bodies through contact},
  author={Posa, Michael and Cantu, Cecilia and Tedrake, Russ},
  journal={The International Journal of Robotics Research},
  volume={33},
  number={1},
  pages={69--81},
  year={2014},
  doi={10.1177/0278364913506757}
}

@inproceedings{pang2021convex,
  title = {A Convex Quasistatic Time-stepping Scheme for Rigid Multibody Systems with Contact and Friction},
  author = {Pang, Tao and Tedrake, Russ},
  booktitle = {IEEE International Conference on Robotics and Automation},
  pages = {6614--6620},
  year = {2021},
  doi = {10.1109/ICRA48506.2021.9560941},
}

@inproceedings{mora2021pods,
  title = {{PODS}: Policy Optimization via Differentiable Simulation},
  author = {Mora, Miguel Angel Zamora and Peychev, Momchil and Ha, Sehoon and Vechev, Martin and Coros, Stelian},
  booktitle = {International Conference on Machine Learning},
  pages = {7805--7817},
  year = {2021},
}

@article{wiedemann2023apg,
  title = {Training Efficient Controllers via Analytic Policy Gradient},
  author = {Wiedemann, Nina and W{\"u}est, Valentin and Loquercio, Antonio and M{\"u}ller, Matthias and Floreano, Dario and Scaramuzza, Davide},
  journal = {IEEE International Conference on Robotics and Automation},
  year = {2023},
}

@article{gillen2022leveraging,
  title = {Leveraging Reward Gradients For Reinforcement Learning in Differentiable Physics Simulations},
  author = {Gillen, Sean and Byl, Katie},
  journal = {arXiv:2203.02857},
  year = {2022},
}

@article{amigo2025decoupled,
  title = {First Order Model-Based RL through Decoupled Backpropagation},
  author = {Amigo, Joseph and Khorrambakht, Rooholla and Chane-Sane, Elliot and Mansard, Nicolas and Righetti, Ludovic},
  journal = {arXiv:2509.00215},
  year = {2025},
}

@article{song2024quadruped,
  title = {Learning Quadruped Locomotion Using Differentiable Simulation},
  author = {Song, Yunlong and Kim, Sangbae and Scaramuzza, Davide},
  journal = {arXiv:2403.14864},
  year = {2024},
}

@inproceedings{schwarke2025learningdeployablelocomotioncontrol,
  title = {Learning Deployable Locomotion Control via Differentiable Simulation},
  author = {Schwarke, Clemens and Klemm, Victor and Bagajo, Joshua and Sleiman, Jean-Pierre and Georgiev, Ignat and Tordesillas, Jesus and Hutter, Marco},
  booktitle = {Conference on Robot Learning},
  year = {2025},
}

@article{luo2024residual,
  title = {Residual Policy Learning for Perceptive Quadruped Control Using Differentiable Simulation},
  author = {Luo, Jing Yuan and Song, Yunlong and Klemm, Victor and Shi, Fan and Scaramuzza, Davide and Hutter, Marco},
  journal = {arXiv:2410.03076},
  year = {2024},
}

@inproceedings{geilinger2020add,
  title={{ADD}: Analytically Differentiable Dynamics for Multi-Body Systems with Frictional Contact},
  author={Geilinger, Moritz and Hahn, David and Zehnder, Jonas and B{\"a}cher, Moritz and Thomaszewski, Bernhard and Coros, Stelian},
  booktitle={ACM Transactions on Graphics (SIGGRAPH Asia)},
  year={2020},
  doi={10.1145/3414685.3417766}
}

@inproceedings{deavila2018end,
  title = {End-to-End Differentiable Physics for Learning and Control},
  author = {de Avila Belbute-Peres, Filipe and Smith, Kevin and Allen, Kelsey and Tenenbaum, Josh and Kolter, J. Zico},
  booktitle = {Advances in Neural Information Processing Systems},
  year = {2018},
}

@article{degrave2019differentiable,
  title={A Differentiable Physics Engine for Deep Learning in Robotics},
  author={Degrave, Jonas and Hermans, Michiel and Dambre, Joni and wyffels, Francis},
  journal={Frontiers in Neurorobotics},
  volume={13},
  pages={6},
  year={2019},
  doi={10.3389/fnbot.2019.00006}
}

@inproceedings{qiao2021efficient,
  title = {Efficient Differentiable Simulation of Articulated Bodies},
  author = {Qiao, Yi-Ling and Liang, Junbang and Koltun, Vladlen and Lin, Ming C.},
  booktitle = {International Conference on Machine Learning},
  year = {2021},
}

@inproceedings{budhiraja2018differential,
  title = {Differential Dynamic Programming for Multi-Phase Rigid Contact Dynamics},
  author = {Budhiraja, Rohan and Carpentier, Justin and Mastalli, Carlos and Mansard, Nicolas},
  booktitle = {IEEE-RAS International Conference on Humanoid Robots},
  pages = {1--9},
  year = {2018},
  doi = {10.1109/HUMANOIDS.2018.8624925},
}

@article{peng2018deepmimic,
  title={{DeepMimic}: Example-Guided Deep Reinforcement Learning of Physics-Based Character Skills},
  author={Peng, Xue Bin and Abbeel, Pieter and Levine, Sergey and van de Panne, Michiel},
  journal={ACM Transactions on Graphics},
  volume={37},
  number={4},
  pages={143:1--143:14},
  year={2018},
  doi={10.1145/3197517.3201311}
}

@article{hwangbo2019agile,
  title={Learning agile and dynamic motor skills for legged robots},
  author={Hwangbo, Jemin and Lee, Joonho and Dosovitskiy, Alexey and Bellicoso, Dario and Tsounis, Vassilios and Koltun, Vladlen and Hutter, Marco},
  journal={Science Robotics},
  volume={4},
  number={26},
  year={2019},
  doi={10.1126/scirobotics.aau5872}
}

@article{lee2020quadrupedal,
  title={Learning quadrupedal locomotion over challenging terrain},
  author={Lee, Joonho and Hwangbo, Jemin and Wellhausen, Lorenz and Koltun, Vladlen and Hutter, Marco},
  journal={Science Robotics},
  volume={5},
  number={47},
  year={2020},
  doi={10.1126/scirobotics.abc5986}
}

@inproceedings{sferrazza2024humanoidbench,
  title = {{HumanoidBench}: Simulated Humanoid Benchmark for Whole-Body Locomotion and Manipulation},
  author = {Sferrazza, Carmelo and Huang, Dun-Ming and Lin, Xingyu and Lee, Youngwoon and Abbeel, Pieter},
  booktitle = {Robotics: Science and Systems},
  year = {2024},
  doi = {10.15607/RSS.2024.XX.061},
}

@inproceedings{cheng2024parkour,
  title = {Extreme Parkour with Legged Robots},
  author = {Cheng, Xuxin and Shi, Kexin and Agarwal, Ananye and Pathak, Deepak},
  booktitle = {IEEE International Conference on Robotics and Automation},
  year = {2024},
}

@INPROCEEDINGS{handa2023dextreme,
  author = {Handa, Ankur and Allshire, Arthur and Makoviychuk, Viktor and Petrenko, Aleksei and Singh, Ritvik and Liu, Jingzhou and Makoviichuk, Denys and Van Wyk, Karl and Zhurkevich, Alexander and Sundaralingam, Balakumar and Narang, Yashraj},
  booktitle = {IEEE International Conference on Robotics and Automation},
  title = {DeXtreme: Transfer of Agile In-hand Manipulation from Simulation to Reality},
  year = {2023},
  volume = {},
  number = {},
  pages = {5977-5984},
  doi = {10.1109/ICRA48891.2023.10160216},
}

@InProceedings{pmlr-v164-rudin22a,
  title = {Learning to Walk in Minutes Using Massively Parallel Deep Reinforcement Learning},
  author = {Rudin, Nikita and Hoeller, David and Reist, Philipp and Hutter, Marco},
  booktitle = {Conference on Robot Learning},
  pages = {91--100},
  year = {2022},
  publisher = {PMLR},
}

@article{doi:10.1126/scirobotics.adi9579,
  author = {Ilija Radosavovic  and Tete Xiao  and Bike Zhang  and Trevor Darrell  and Jitendra Malik  and Koushil Sreenath },
  title = {Real-world humanoid locomotion with reinforcement learning},
  journal = {Science Robotics},
  volume = {9},
  number = {89},
  pages = {eadi9579},
  year = {2024},
  doi = {10.1126/scirobotics.adi9579},
}

@inproceedings{werling2021fast,
  title = {Fast and Feature-Complete Differentiable Physics Engine for Articulated Rigid Bodies with Contact Constraints},
  author = {Werling, Keenon and Omens, Dalton and Lee, Jeongseok and Exarchos, Ioannis and Liu, C. Karen},
  booktitle = {Robotics: Science and Systems},
  year = {2021},
  doi = {10.15607/RSS.2021.XVII.034},
}

@article{howell2022dojo,
  title = {{Dojo}: A Differentiable Simulator for Robotics},
  author = {Howell, Taylor A. and Le Cleac'h, Simon and Kolter, J. Zico and Schwager, Mac and Manchester, Zachary},
  journal = {arXiv:2203.00806},
  year = {2022},
}

@Inbook{Moreau1988,
  author = "Moreau, J. J.",
  title = "Unilateral Contact and Dry Friction in Finite Freedom Dynamics",
  bookTitle = {Nonsmooth Mechanics and Applications},
  year = "1988",
  publisher = "Springer Vienna",
  pages = "1--82",
  doi = "10.1007/978-3-7091-2624-0_1",
}

@article{harvey2020robust,
  author = {F{\'e}lix G. Harvey and Mike Yurick and Derek Nowrouzezahrai and Christopher Pal},
  title = {Robust Motion In-Betweening},
  journal = {ACM Transactions on Graphics},
  volume = {39},
  number = {4},
  year = {2020},
  publisher = {ACM}
}

\end{document}